\documentclass[11pt]{article}

\usepackage[final]{acl}

\usepackage{microtype}
\usepackage{graphicx}
\usepackage{booktabs}
\usepackage{amsmath}
\usepackage{amssymb}
\usepackage[english,bidi=default]{babel} 
\babelfont{rm}{TeXGyreTermesX} 
\babelprovide[import]{hindi}
\babelfont[*devanagari]{rm}{Lohit Devanagari}
\babelprovide[import]{arabic}
\babelfont[*arabic]{rm}{Noto Sans Arabic}

\title{SemBlock: Semantic Boundary Dynamic Blocks for Diffusion LLMs}

\author{
Xinrui Song$^{*1}$,
Zhuoran Wang$^{*1}$,
Mingju Gao$^{1}$,
Hao Tang$^{\dagger1}$ \\
\\[-0.3em]
$^{1}$School of Computer Science, Peking University \\
\\[-0.3em]
\texttt{202300800600@mail.sdu.edu.cn, haotang@pku.edu.cn}
}

\begin{document}

\maketitle

\begingroup
\renewcommand\thefootnote{}
\footnotetext{
$^*$Equal contribution. 
$^\dagger$Corresponding author.
}
\endgroup

\begin{abstract}

Diffusion language models (DLMs) generate text through iterative denoising, and blockwise decoding improves their practicality by committing tokens in local blocks. However, existing blockwise methods typically rely on fixed block sizes or delimiter-based runtime signals, which do not necessarily align with semantic boundaries. In this paper, we propose \textbf{SemBlock}, a semantic-boundary-driven dynamic block decoding framework for diffusion LLMs. SemBlock formulates dynamic block construction as semantic boundary prediction and trains lightweight predictors on frozen LLaDA hidden states. To provide supervision, we construct \textbf{SemBound}, a semantic-boundary dataset that derives boundary labels from discourse units, reasoning steps, and implementation spans across natural language, math, and code tasks. During inference, SemBlock uses predicted boundary probabilities to select the ending position of each dynamic block. Experiments on GSM8K, IFEval, MATH, and HumanEval show that SemBlock consistently improves over fixed-block decoding and AdaBlock. Our code is publicly available: \href{https://github.com/TH-AI-Lab-PKU/SemBlock}{https://github.com/TH-AI-Lab-PKU/SemBlock}.
\end{abstract}




\section{Introduction}

\begin{figure*}[t]
    \centering
    \includegraphics[
        width=\textwidth,
        trim=0.0in 3.9in 0.0in 3.9in,
        clip
    ]{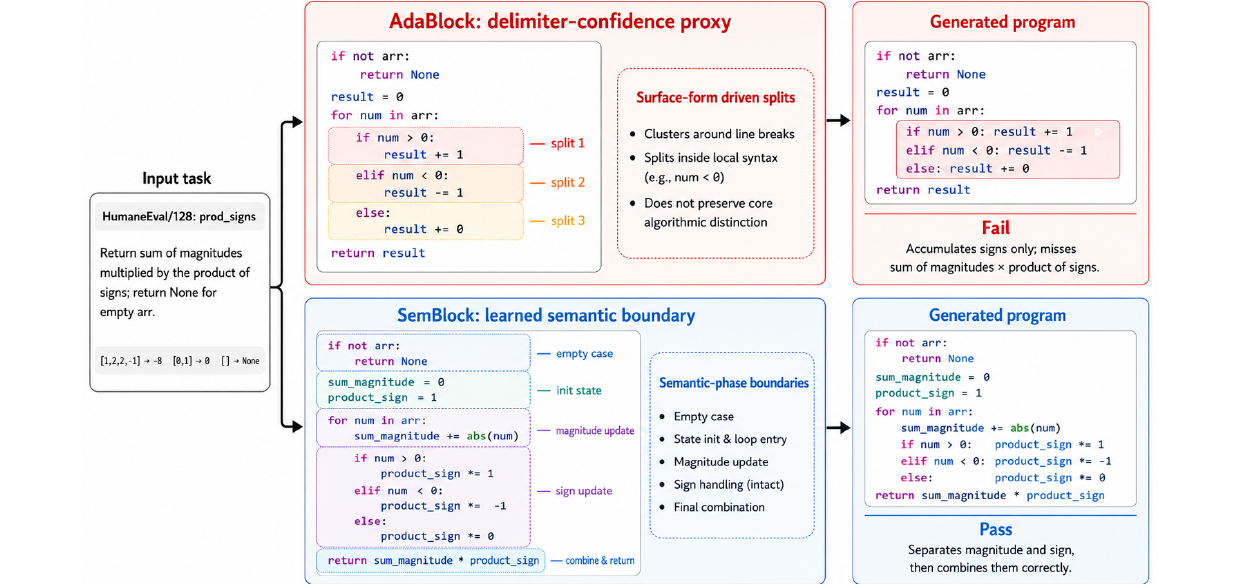}
    \caption{
    Boundary comparison on HumanEval/128 \texttt{prod\_signs}. 
    AdaBlock follows delimiter-confidence cues and produces a surface-driven segmentation, while SemBlock aligns boundaries with semantic implementation phases and generates the correct program.
    }
    \label{fig:adablock_semblock_case}
\end{figure*}

Diffusion language models (DLMs) generate text through iterative denoising over masked token sequences, and blockwise decoding makes
this paradigm practical by committing tokens in local blocks rather than refining the entire output at once~\cite{li2022diffusionlm,lou2024discrete,sahoo2024simple,nie2025large,arriola2025blockdiffusion,hersche2026soft,jazbec2026learning}.
The key design question in blockwise decoding is where each block should end: the boundary determines which tokens are denoised and committed together. Fixed block sizes are simple but introduce a semantic mismatch — a block that ends too early commits a fragmented output, while one that extends
  too far includes tokens not yet reliably constrained by context~\citep{lu2025adablock,luo2026dsb,xia2026tstar}.
  AdaBlock addresses this by adjusting block length at runtime via
  delimiter tokens and their denoising confidence scores~\citep{lu2025adablock},
  but delimiter-based signals remain an indirect proxy: the candidate cuts
  follow surface markers such as punctuation and line breaks, which do not
  reliably coincide with semantic unit completion across natural language,
  mathematical reasoning, or code generation~\citep{zeldes2017gum,zeldes2019disrpt,yu2019gumdrop,ling2017program,wei2022chain,cobbe2021training,chen2021evaluating}.
  As shown in Figure~\ref{fig:adablock_semblock_case}, this leads to
  boundaries that cut across functional implementation phases rather than
  respecting program structure.

We propose \textbf{SemBlock}, which formulates dynamic block
  construction as a semantic boundary prediction problem.
  A lightweight predictor built on frozen LLaDA hidden
  states~\citep{nie2025large} estimates, at each candidate position,
  whether the current semantic unit has been completed.
  To provide training supervision, we construct \textbf{SemBound}, a
  semantic boundary dataset that derives boundary labels from discourse
  units in natural language~\citep{zeldes2017gum,zeldes2019disrpt,yu2019gumdrop}, reasoning steps in mathematical solutions~\citep{ling2017program,wei2022chain,cobbe2021training},
  and implementation spans in code~\citep{husain2019codesearchnet,li2022alphacode}.
  During inference, the predictor outputs boundary probabilities over the
  candidate block window, and the scheduler selects the most confident
  position above a threshold as the block endpoint; if no position clears
  the threshold, decoding falls back to the default block size.

  Experiments on GSM8K~\cite{cobbe2021training},
  IFEval~\cite{zhou2023instructionfollowing},
  MATH~\cite{hendrycks2021measuring}, and
  HumanEval~\cite{chen2021evaluating} show that SemBlock consistently
  outperforms fixed-block decoding and AdaBlock under the same initial
  block budget.
  On LLaDA-1.5~\citep{zhu2025llada15}, SemBlock further improves over AdaBlock by up to 11.60
  pass@1 points on HumanEval.
  Ablation studies confirm that the gain does not come from finer
  segmentation or more denoising computation, but from placing block
  boundaries closer to true semantic units.

  Our contributions are as follows:
  \begin{itemize}
      \item We propose \textbf{SemBlock}, a semantic-boundary-driven
      dynamic block decoding framework for diffusion LLMs that replaces
      runtime delimiter-confidence proxies with learned semantic boundary
      prediction.
      \item We construct \textbf{SemBound}, a semantic boundary
      supervision dataset covering natural language, mathematical
      reasoning, and code generation, with labels derived from discourse
      units, reasoning steps, and implementation spans.
      \item Experiments on four benchmarks show that SemBlock improves
      over fixed-block decoding and AdaBlock, with ablations confirming
      that the gain comes from more accurate semantic boundary placement.
  \end{itemize}

\section{Related Work}

\subsection{DLMs and Blockwise Decoding}

DLMs generate text through iterative denoising rather than following the strictly left-to-right generation order used by conventional autoregressive models \citep{li2022diffusionlm,lou2024discrete,sahoo2024simple}. Recent large-scale masked diffusion models, such as LLaDA and Dream, have shown that diffusion-based generation can serve as a competitive alternative to autoregressive large language models while naturally supporting parallel token refinement. More recent studies further extend this line of work to soft masking, language-specific masked diffusion models, latent-space diffusion language modeling, and speculative decoding for masked diffusion models \citep{hersche2026soft}.

Since fully non-autoregressive decoding often struggles to capture strong sequential dependencies, recent DLMs commonly adopt a blockwise semi-autoregressive paradigm: tokens within the same block are denoised in parallel, whereas different blocks are finalized sequentially \citep{arriola2025blockdiffusion,wu2025fastdllm}. This design preserves part of the dependency structure of autoregressive decoding while enabling block-level KV caching, making it a practical decoding strategy for efficient diffusion language model inference.

\subsection{Block Scheduling in DLMs}

Blockwise DLM decoding commonly relies on a predefined default block size. This setting is simple and stable, but once the block size is fixed, the sampler must make decoding decisions within the current block~\cite{swordsman2026}.
AdaBlock systematically questions this fixed-block assumption. It shows that fixed blocks may delay the decoding of high-confidence tokens outside the current block and may also force low-confidence tokens inside the current block to be committed too early. To mitigate this issue, AdaBlock analyzes confidence dynamics during denoising~\cite{ghazvininejad2019maskpredict} and adjusts block length at runtime using delimiter tokens and their confidence scores.

Such a proxy can be effective when delimiters align well with semantic steps, but this alignment is not always stable across natural language, mathematical reasoning, and code generation. In natural language, semantic boundaries are closer to discourse units than to punctuation or line breaks alone~\citep{zeldes2017gum,zeldes2019disrpt,yu2019gumdrop}. In mathematical reasoning, boundaries often correspond to the completion of reasoning states or intermediate steps~\citep{ling2017program,wei2022chain,cobbe2021training}. In code generation, useful boundaries are also tied to program structure and implementation intent, especially when functional correctness is evaluated through executable programs~\citep{chen2021evaluating}.
However, the boundary signal in AdaBlock is still derived from a runtime delimiter-confidence proxy~\citep{lu2025adablock}. It indirectly approximates the end of a semantic step through delimiters and their confidence scores, rather than directly predicting semantic boundaries themselves.


\subsection{Semantic Boundaries for Structured Generation}

Semantic boundaries have been studied in different forms across natural language, reasoning, and code. Discourse segmentation divides text into discourse units \citep{zeldes2017gum,zeldes2019disrpt,yu2019gumdrop}; reasoning-step decomposition describes intermediate reasoning states \citep{wei2022chain,cobbe2021training}; and program-structure modeling captures code spans and functional components \citep{ling2017program}.

These studies usually treat segmentation as the task output. In contrast, our goal is not to recover discourse units, reasoning steps, or program spans for their own sake, but to use semantic boundaries as control signals for dynamic block decoding in DLMs. When these predictions are sufficiently aligned with semantic transitions, the resulting workflow improves over AdaBlock and achieves stronger performance than the compared adaptive decoding methods.


\section{SemBound Dataset}
\label{sec:data}

\begin{figure*}[t]
    \centering
    \includegraphics[width=\textwidth]{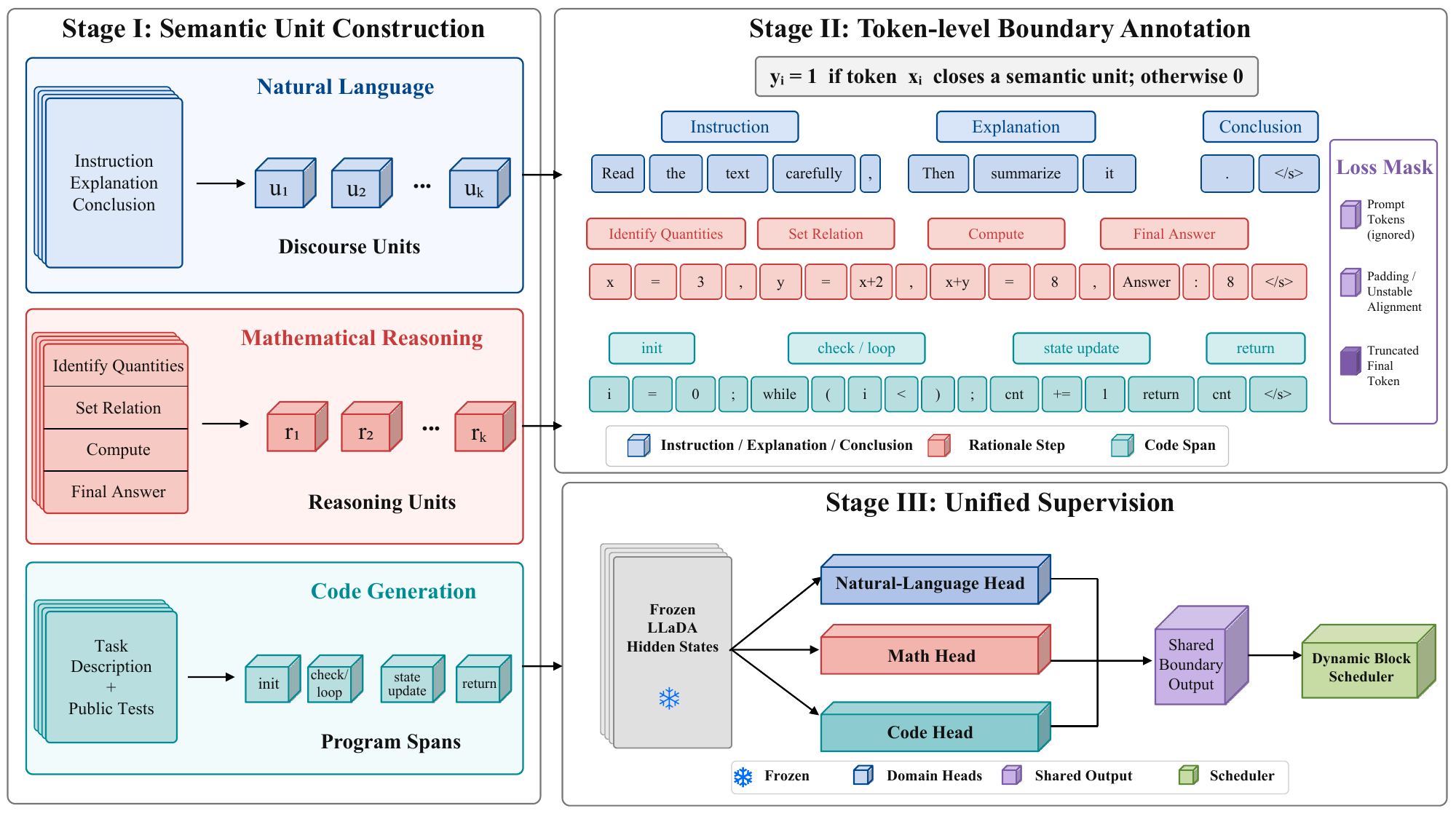}
    \caption{
    Overview of the SemBound data construction pipeline.
    Natural language, mathematical, and code generation examples are organized into domain specific semantic units and converted into token-level boundary labels for unified boundary supervision.
    }
    \label{fig:sembound_pipeline}
\end{figure*}


\textbf{Motivation.} We broadly categorize generation tasks for diffusion language models into three major types: natural-language generation, mathematical reasoning, and code generation. For blockwise decoding, all three settings require the scheduler to decide where the next block should end. This decision is not merely a length-control problem: the boundary determines which tokens are refined and committed together. 
This motivates the construction of supervision that indicates when a semantic unit has been completed. Delimiters, punctuation, and line breaks can provide local cues, but they are not equivalent to semantic closure itself.

We therefore construct SemBound to provide semantic boundary supervision for SemBlock. The goal of SemBound is to convert semantic completion positions in different domains into token-level training signals that can be used by the dynamic block scheduler.

Figure~\ref{fig:semantic_boundary_framework} illustrates the construction pipeline of SemBound. The pipeline contains three stages: semantic unit construction, token-level boundary annotation, and unified supervision for the boundary predictor.

\subsection{Semantic Unit Construction}

In Stage I, each example is first organized into an ordered sequence of semantic units. Natural-language examples are divided into discourse units, mathematical examples into reasoning units, and code examples into program spans. These units come from different domain structures, but they share the same role: each unit represents a locally complete span that can be generated and committed as a whole.

\noindent\textbf{Natural-language Generation.}
  For natural language data, we derive segment boundaries from GUM
  discourse segmentation annotations. Each annotated segment corresponds
  to a contiguous discourse unit with a functional role such as
  instruction, explanation, or conclusion. The final token of each segment
  is labeled as a boundary position, and all target tokens contribute to
  the boundary loss.


\noindent\textbf{Mathematical Reasoning.}
  For mathematical data, we derive segments from solution rationales in
  AQuA-RAT. The rationale is split and merged into units that each
  correspond to a complete reasoning function: identifying quantities,
  setting up relations, computing, or producing the final answer.
  A boundary is placed at the final token of each unit.

\noindent\textbf{Code Generation.}
  For code data, we organize each example as a function generation task,
  as shown in Figure~\ref{fig:sembound_pipeline}. The conditioning context
  consists of the task description, function signature, and, when
  available, compact public test information; boundary supervision is
  applied only to the target code region. Program-structure analysis
  annotates the target region with three types of token-level labels:
  \emph{phase labels} that identify the implementation role of each span; \emph{transition labels} that indicate whether a phase
  change occurs after the current token; and \emph{boundary-type labels}
  that describe the nature of each transition. These three signals provide
  auxiliary supervision for the code head.

\subsection{Token Level Boundary Annotation and Unified Supervision}
In Stage II, segment-level semantic units are converted into token-level boundary labels. The final token of each segment is marked as a boundary, indicating that a semantic unit is completed after this position. Other tokens inside the segment are marked as non-boundary positions. The resulting token-level labels are used as training targets for the boundary predictor.

For examples with prompts or conditioning information, prompt tokens are kept as context but are not used as boundary-loss targets. Padding positions, unstable alignment positions, and truncated final tokens are also excluded by the loss mask, so that unreliable positions do not become boundary supervision.
All domains are finally converted into a unified training representation:
\[
\mathcal{D} =
\left\{
x_{1:n},\;
y_{1:n},\;
m_{1:n},\;
d
\right\},
\]
where $d$ denotes the domain of the example. 
For code examples, the record additionally contains phase labels, transition labels, boundary type labels, and their validity masks. 
For natural-language and mathematical examples, these structural fields are inactive.

In Stage III, token-level boundary labels from different domains are unified into a common training target. Frozen LLaDA hidden states provide contextual representations, while the natural-language head, math head, and code head learn boundary patterns in their corresponding domains. All heads produce boundary probabilities in the same format, which are then used by the dynamic block scheduler.

Through this construction, SemBound converts domain-specific semantic units into unified token-level boundary supervision. This allows SemBlock to learn when the current block should end, rather than learning where punctuation or delimiters appear.

\section{Methodology}

\subsection{Overview}


\begin{figure*}[t]
    \centering
    \includegraphics[width=\textwidth]{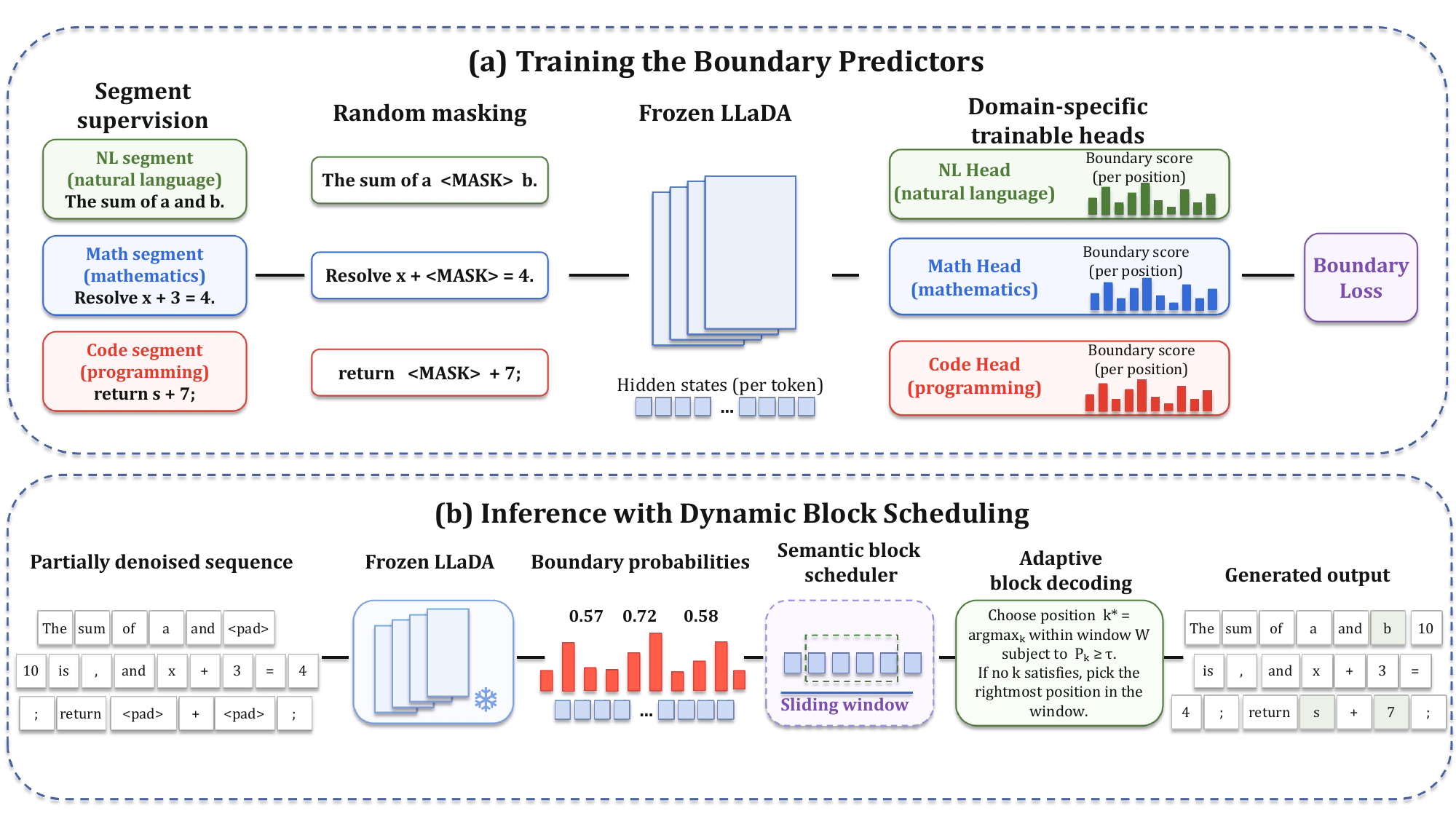}
    \caption{
    Semantic boundary guided dynamic block prediction. 
    The frozen LLaDA backbone provides hidden states, while trainable boundary heads predict semantic boundary probabilities for dynamic block scheduling.
    }
    \label{fig:semantic_boundary_framework}
\end{figure*}


We propose SemBlock, a semantic boundary prediction framework for dynamic block decoding in diffusion language models. As shown in Figure~\ref{fig:semantic_boundary_framework}, supervision from different domains is first organized into semantic segments. These segments correspond to discourse units in natural language, reasoning steps in mathematical tasks, and program-level spans in code generation. We then derive token-level boundary labels by marking the final token of each segment as a boundary, which provides a common training target for the boundary predictor.


The overall training and inference workflow is shown in Figure~\ref{fig:semantic_boundary_framework}. During training, these token-level boundary labels supervise lightweight boundary heads built on top of frozen LLaDA hidden states. Given a partially observed denoising state containing both visible and masked tokens, the frozen LLaDA backbone produces contextual hidden representations. The boundary predictor then estimates, for each token position, the probability that the transition after this token closes a semantic unit. During inference, these probabilities serve as the scheduling signal for dynamic block construction, allowing the next block boundary to be selected according to the learned semantic structure of the current sequence.

Formally, let $\mathbf{x} = (x_1,\ldots,x_n)$ denote the current token sequence, and let $\mathbf{h}_i \in \mathbb{R}^d$ be the hidden state at position $i$ produced by the frozen LLaDA model. The boundary predictor estimates
\[
p_i = P(y_i = 1 \mid \mathbf{h}_1,\ldots,\mathbf{h}_n),
\]
where $y_i=1$ indicates that a semantic boundary occurs immediately after token $x_i$. The resulting boundary probabilities are then used to guide the construction of the next decoding block.

\subsection{Semantic Boundary Prediction}
\label{sec:boundary_prediction}

\textbf{Architecture.}
We build a lightweight semantic boundary predictor on top of a frozen LLaDA backbone. Given a partially observed denoising state, LLaDA produces contextual hidden states for both visible and masked tokens. These hidden states are fed into a trainable boundary module, which first constructs shared boundary representations and then routes them to domain-specific heads. The domain-specific heads capture different forms of semantic closure. The natural-language head predicts discourse or instruction boundaries, the mathematical head predicts transitions between reasoning steps, and the code head predicts structural or intent boundaries in program generation. Although these heads use different supervision signals, they share the same output format: token-level boundary probabilities used for dynamic block scheduling.

We formulate boundary prediction as transition classification. The goal is to predict whether the position after token $x_i$ corresponds to the end of a semantic segment. Instead of using only the hidden state at position $i$, the predictor also incorporates the hidden state at the adjacent position $i+1$, so that the boundary decision can be conditioned on the local change in contextual representations.
\noindent\textbf{Training.}
During training, the boundary predictor defined above is the only component being optimized. 
The objective is to identify reliable semantic closure points under partially observed denoising states. 
To approximate the input condition encountered during diffusion decoding, we randomly mask a subset of valid target tokens in each example before computing the boundary predictions.

Examples from natural language, mathematical reasoning, and code generation use their corresponding supervision signals. 
For each training example, the loss is applied only to the prediction head associated with its domain, while the shared boundary module receives gradients from all domains. 
This keeps the output interface unified while allowing each domain to preserve its own boundary criterion.

Let $\ell_i$ denote the boundary logit at position $i$, and let $y_i \in \{0,1\}$ denote the boundary label. 
We use a masked binary classification objective and compute the loss only on target positions with reliable boundary supervision. 
Prompt tokens, padding tokens, and positions without reliable annotations are excluded. 
Since boundary labels are sparse, positive boundary positions are assigned larger class weights:
\[
\mathcal{L}_{\text{bdry}}
=
\frac{
\sum_i m_i \, w_i \, \ell_{\text{BCE}}(\ell_i, y_i)
}{
\sum_i m_i \, w_i
},
\]
where $m_i$ is the loss mask and $w_i$ is the class weight.

For code generation, a boundary often corresponds not only to a local pause in the token sequence, but also to a transition between functional units of the program. 
For example, variable initialization, local computation, and return generation play different structural roles. 
The code head therefore uses additional structural signals, including program phase, phase transition, and boundary type, so that boundary prediction can reflect functional changes during program generation.

\noindent\textbf{Inference.}
During inference, the boundary predictor outputs boundary probabilities for positions inside the next candidate block window. 
The scheduler does not modify the denoising objective of the base model; it only uses the predicted semantic boundary signal to determine the current block length. 
Given the default block size $B$ and threshold $\tau$, the scheduler selects the most confident boundary position whose probability exceeds the threshold:
\[
b^\star =
\arg\max_{b \in [1,B]} p_b
\quad
\text{s.t.}
\quad p_b \ge \tau .
\]

If no position in the window reaches the threshold, the scheduler keeps the default block size $B$. 
Thus, SemBlock changes the block length only when the boundary prediction is sufficiently reliable. 
When the signal is uncertain, it falls back to fixed-block decoding and avoids unstable over-segmentation.

\section{Experiments}

\subsection{Experimental Setup}
\label{sec:experimental_setup}

\textbf{Models and Hardware.}
We evaluate SemBlock on representative diffusion large language models, including LLaDA-Instruct and LLaDA-1.5~\citep{zhu2025llada15}. Unless otherwise stated, the main experiments use LLaDA-Instruct and are conducted on a single NVIDIA H20 GPU. Detailed decoding hyperparameters and task settings are provided in Appendix~\ref{app:additional_details}.



\noindent\textbf{Metrics.}
Since the evaluated tasks differ in their output format and correctness criteria, we report the standard metric for each benchmark. For mathematical reasoning, we report answer accuracy, which measures whether the final answer matches the reference answer. For IFEval, we report prompt-level strict accuracy, which evaluates whether the generated response satisfies all required instruction constraints. For HumanEval, we report execution-based functional correctness. Specifically, pass@1 measures the percentage of tasks for which the single generated completion passes the unit tests.

\noindent\textbf{Baselines.}
We compare SemBlock with Vanilla LLaDA, Fixed Block, AdaBlock, and AdaBlock + Cache. Vanilla LLaDA denotes the base decoding setting. Fixed Block keeps the default block size unchanged. AdaBlock uses adaptive block scheduling. AdaBlock + Cache further adds cache to AdaBlock. SemBlock denotes our method.

\subsection{Main Results}

\subsubsection{Generation Quality on LLaDA-Instruct}
Table~\ref{tab:main_results_b32} reports the main results at $B_0=32$. This setting directly compares SemBlock with Fixed Block, AdaBlock, and AdaBlock + Cache. 

\begin{table*}[t]
\centering
\small
\setlength{\tabcolsep}{5.5pt}
\begin{tabular}{lccccccc}
\toprule
Task & Metric & Vanilla LLaDA & Fixed Block & AdaBlock & AdaBlock + Cache & SemBlock & Gain \\
\midrule
GSM8K & Acc. & 76.70 & 77.60 & 80.60 & 78.50 & \textbf{82.60} & \textbf{+2.00} \\
IFEval & Acc. & -- & 55.64 & 55.45 & -- & \textbf{56.93} & \textbf{+1.29} \\
MATH & Acc. & 36.90 & 36.90 & 37.30 & 35.30 & \textbf{37.80} & \textbf{+0.50} \\
HumanEval & pass@1 & 43.90 & 43.90 & 43.30 & 46.30 & \textbf{46.95} & \textbf{+0.65} \\
\bottomrule
\end{tabular}
\caption{
Main results on LLaDA-Instruct at $B_0=32$. GSM8K and MATH use answer accuracy, IFEval uses prompt-level strict accuracy, and HumanEval uses pass@1. Gain denotes the improvement of SemBlock over the strongest non-SemBlock method in each row.
}
\label{tab:main_results_b32}
\end{table*}

\noindent\textbf{Generation Quality Across Tasks.}
Table~\ref{tab:main_results_b32} reports the main results at $B_0=32$. SemBlock achieves the best result on all four benchmarks. It improves GSM8K from 80.60 to 82.60 over AdaBlock, IFEval from 55.45 to 56.93 over AdaBlock, MATH from 37.30 to 37.80 over AdaBlock, and HumanEval from 46.30 to 46.95 over the strongest non-SemBlock baseline. These results show that learned semantic boundaries improve dynamic block decoding across instruction following, mathematical reasoning, and code generation.

\noindent\textbf{Improvement Over Fixed and Adaptive Block Baselines.}
Compared with Fixed Block, SemBlock improves GSM8K by 5.00 points, IFEval by 1.29 points, and MATH by 0.90 points. Compared with AdaBlock, SemBlock improves GSM8K by 2.00 points, IFEval by 1.48 points, and MATH by 0.50 points. These gains indicate that supervised semantic boundaries provide a stronger scheduling signal than either a fixed window or adaptive boundaries derived from surface markers.

\noindent\textbf{Code Generation.}
On HumanEval, SemBlock reaches 46.95 pass@1, improving over AdaBlock + Cache by 0.65 points and over AdaBlock by 3.65 points. Since pass@1 measures whether a generated completion passes the unit tests, the improvement suggests that learned program boundaries help produce more functionally correct code.

\subsubsection{Generation Quality on LLaDA-1.5}

\begin{table}[t]
\centering
\normalsize
\setlength{\tabcolsep}{8pt}
\renewcommand{\arraystretch}{1.12}
\begin{tabular}{lccc}
\toprule
Method & $B_0=16$ & $B_0=32$ & $B_0=64$ \\
\midrule
AdaBlock & 37.80 & 38.40 & 38.40 \\
SemBlock & \textbf{47.56} & \textbf{49.39} & \textbf{50.00} \\
$\Delta$ & \textbf{+9.76} & \textbf{+10.99} & \textbf{+11.60} \\
\bottomrule
\end{tabular}
\caption{
HumanEval results on LLaDA-1.5. $\Delta$ denotes the improvement of SemBlock over AdaBlock, measured by pass@1.
}
\label{tab:humaneval_llada15_block_size}
\end{table}

Table~\ref{tab:humaneval_llada15_block_size} further evaluates SemBlock on HumanEval with LLaDA-1.5. SemBlock consistently outperforms AdaBlock across all default block sizes, improving pass@1 by 9.76, 10.99, and 11.60 points at $B_0=16,32,64$, respectively.
While AdaBlock remains nearly unchanged across block sizes, SemBlock improves from 47.56 at $B_0=16$ to 50.00 at $B_0=64$. This trend suggests that learned code boundaries can better exploit a larger candidate window, where the scheduler has more room to select a suitable program boundary.


\subsection{Ablation Studies}
\label{sec:ablation}

We conduct two ablation studies to examine the role of semantic boundaries in SemBlock. The first study asks whether SemBlock simply reproduces AdaBlock boundaries. The second study asks whether finer segmentation or more computation necessarily leads to better performance. Together, these studies show that the key factor in dynamic block prediction is not merely changing the block size, but choosing boundaries that match semantic units.

\subsubsection{Boundary Overlap with AdaBlock}

We therefore compare the boundary sets selected by SemBlock and AdaBlock on the same samples. Exact overlap measures the percentage of SemBlock boundaries that are matched by AdaBlock at the same position. Jaccard overlap measures the intersection over union between the two boundary sets. A lower Jaccard score indicates a larger difference in the overall segmentation structure.

\begin{table}[t]
\centering
\small
\setlength{\tabcolsep}{4.5pt}
\renewcommand{\arraystretch}{1.08}
\begin{tabular}{llccc}
\toprule
Task & Head & Samples & Exact & Jaccard \\
\midrule
HumanEval & Code & 17 & 4.76 & 1.95 \\
IFEval & GUM & 55 & 25.88 & 11.25 \\
GSM8K & Math & 132 & 65.45 & 51.35 \\
MATH & Math & 351 & 82.39 & 69.19 \\
\bottomrule
\end{tabular}
\caption{
Boundary overlap between SemBlock and AdaBlock. Exact and Jaccard are reported as percentages. MATH uses 351 samples.
}
\label{tab:boundary_overlap_adablock}
\end{table}

Table~\ref{tab:boundary_overlap_adablock} shows that the overlap is very low on HumanEval and IFEval. The Jaccard score is only 1.95 on HumanEval and 11.25 on IFEval. This indicates that the code head and the natural-language head do not learn AdaBlock-style surface boundaries, but instead capture program structure and discourse structure.

The overlap is higher on math tasks, but this does not mean that the two methods are equivalent. In GSM8K and MATH, line breaks, equations, and reasoning steps can naturally align with delimiter cues, so AdaBlock and SemBlock may agree on some positions. However, the Jaccard score is still only 51.35 on GSM8K, and MATH is also far from perfect agreement. Thus, even on math tasks where surface cues are most likely to align with reasoning steps, SemBlock is not a direct copy of AdaBlock.

This provides strong evidence that boundaries derived only from local confidence or surface markers are different from boundaries learned through semantic supervision.

\subsubsection{Effect of Boundary Signals on GSM8K}

However, different boundaries do not necessarily mean better boundaries. We therefore conduct a controlled ablation on GSM8K to test whether different boundary signals affect generation quality. The model, task, default block size, generation length, number of denoising steps, and cache setting are kept fixed; only the boundary signal is changed.

We compare three boundary settings. Hybrid uses both the mathematical semantic boundary head and the natural-language delimiter cue. Delimiter only keeps only the natural-language delimiter cue, AdaBlock-style surface boundary. Math head only keeps only the mathematical semantic boundary head and removes the influence of delimiter confidence.

\begin{table}[t]
\centering
\small
\setlength{\tabcolsep}{4.8pt}
\renewcommand{\arraystretch}{1.08}
\begin{tabular}{lccc}
\toprule
Variant ($\Delta$) & Strict EM & NFE & Block Len. \\
\midrule
\textbf{Ours } & \textbf{85.33} & \textbf{104.48} & \textbf{19.06} \\
Delimiter only (-3.00) & 82.33 & 102.75 & 20.53 \\
Math head only (-1.67) & 83.67 & 112.08 & 14.66 \\
\bottomrule
\end{tabular}
\caption{
Ablation of boundary signals on GSM8K. All results use the same $B_0=32$ setting and 300 samples. The value in parentheses, $\Delta$, denotes the change in strict exact match relative to Ours. NFE denotes the average number of denoising model forward calls per generated sample.
}
\label{tab:gsm8k_boundary_ablation}
\end{table}

Table~\ref{tab:gsm8k_boundary_ablation} shows that Ours achieves the best strict EM of 85.33. When only the delimiter cue is used, the score drops to 82.33, a decrease of 3.00 points. This shows that relying only on surface delimiters or local confidence signals is insufficient for mathematical reasoning generation.

When only the mathematical semantic boundary head is used, the score is 83.67, still lower than Ours. This indicates that mathematical semantic boundaries and natural-language step cues are complementary.

More importantly, Math head only has an average block length of 14.66, much smaller than 19.06 for Ours, and its average NFE is 112.08, higher than 104.48 for Ours. In other words, Math head only creates finer segments and uses more denoising calls, but still obtains lower accuracy.

This result further challenges the assumption that smaller blocks or more computation necessarily lead to better performance. Overly fine segmentation can break the semantic continuity within a complete reasoning step, causing the model to commit to local fragments too early and thereby degrading the quality of the final answer.

Therefore, Dynamic block decoding achieves the best quality when block boundaries align with both mathematical reasoning steps and explicit step structure in natural language.

Together, the two ablations support our central experimental claim. SemBlock boundaries are different from AdaBlock-style surface boundaries. Relying only on surface signals hurts reasoning quality, and simply making the segmentation finer does not improve performance. The closer the block boundaries are to true semantic units, the better the generation quality of the diffusion language model.

\section{Conclusion}

This work introduces SemBlock, which formulates dynamic block prediction in DLMs as semantic boundary prediction. Built on frozen LLaDA hidden states, SemBlock learns semantic closing points for natural language, mathematical reasoning, and code generation, and uses them to guide dynamic block decoding. Unlike fixed block decoding or schedules based on surface markers, SemBlock aligns block boundaries more closely with semantic units and improves generation quality. Further ablations show that the gain does not come from finer segmentation or more denoising computation, but from more accurate semantic boundary selection. We hope this semantic-boundary perspective will inspire future training and inference strategies for DLMs.

\section{Limitations}

\textbf{Limitations for Boundary Supervision Quality.}
SemBlock introduces semantic boundaries into dynamic block scheduling for DLMs, but the current method still depends on the available boundary supervision. The boundaries used for natural language, mathematical reasoning, and code generation are derived from discourse annotations, rationales, and program structures. These signals provide useful approximations of semantic closure, but they are not equivalent to a unified and manually calibrated annotation of semantic boundaries. When the constructed segments are noisy or incomplete, the boundary predictor may inherit these errors and produce suboptimal scheduling decisions.

This limitation becomes more visible when extending the method to broader generation settings. The three domains studied in this work all provide relatively clear structural cues: discourse units in natural language, reasoning steps in mathematical rationales, and implementation spans in code. In open-ended dialogue, long-document generation, multi-turn tool use, or more complex program synthesis, semantic closure may be more implicit and may not be directly recoverable from existing data structures. Applying SemBlock to these settings will require boundary construction methods that are less tied to a specific dataset format.

\noindent\textbf{Limitations for Frozen backbone.}
SemBlock also keeps the LLaDA backbone frozen. This design preserves the original generative ability of the diffusion language model and avoids retraining the backbone, but it also means that the boundary head can only read semantic signals already encoded in the hidden states. If the backbone does not represent certain reasoning structures, program structures, or long-range semantic dependencies clearly, a lightweight boundary head may not fully recover the desired boundary signal. Future work may explore joint adaptation of the representation model and the boundary predictor, while carefully controlling the additional training cost and stability issues.

\noindent\textbf{Limitations for Inference.}
At inference time, SemBlock still relies on a default block size and a boundary threshold. The current scheduler selects a boundary only inside the candidate window and only when the predicted probability is sufficiently high. If the window size or threshold is not well matched to the task, the schedule may become overly conservative or produce unstable cuts. Similar to other adaptive block methods, dynamic scheduling is not only an accuracy problem; it may also affect NFE, throughput, and the behavior of approximate caching. The current work focuses mainly on generation quality, leaving a more systematic study of the quality-efficiency trade-off to future work.

\noindent\textbf{Future Work.}
Future work will extend SemBlock along three directions: constructing higher-quality and more transferable semantic boundary supervision, designing schedulers that adapt the window and threshold according to the current generation state, and studying how semantic boundary prediction interacts with caching, parallel decoding, and long-context generation.

\bibliography{custom}

\clearpage
\newpage

\appendix

\section{SemBound Data Sources and Splits}
\label{app:sembound-data-sources}

This section summarizes the data sources and split statistics used to construct the SemBound supervision, as described in Section\ref{sec:data}. As shown in Table~\ref{tab:sembound-data-splits}, SemBound covers three generation scenarios: natural-language generation, mathematical reasoning, and code generation. 
The natural-language and mathematical-reasoning splits are constructed from GUM discourse segmentation and AQuA-RAT rationales, respectively, while the code-generation split is built from mixed function-completion records. 
Table~\ref{tab:sembound-code-source-breakdown} further reports the source composition of the code-generation training split~\cite{husain2019codesearchnet,li2022alphacode}.

\begin{table*}[t]
\centering
\small
\begin{tabular}{llrrr}
\toprule
Scenario & Source & Train & Valid & Test \\
\midrule
Natural language 
& GUM discourse segmentation 
& 211 & 32 & 32 \\
Mathematical reasoning 
& AQuA-RAT (Math external / combined) 
& 97,467 & 254 & 254 \\
Code generation 
& Mixed function-completion records 
& 20,804 & 2,525 & 164 \\
\bottomrule
\end{tabular}
\caption{
Data sources and train/validation/test splits used for SemBound boundary supervision.
Natural-language supervision is constructed from GUM discourse segmentation annotations; mathematical reasoning supervision is constructed from AQuA-RAT rationales under the Math external / combined setting; code supervision is constructed from mixed function-completion records.
}
\label{tab:sembound-data-splits}
\end{table*}

\begin{table}[t]
\centering
\small
\begin{tabular}{lr}
\toprule
Source & Training examples \\
\midrule
CodeSearchNet & 6,554 \\
LeetCode & 10,800 \\
CodeContests & 1,200 \\
Synthetic non-HumanEval & 2,250 \\
\midrule
Total & 20,804 \\
\bottomrule
\end{tabular}
\caption{
Source breakdown of the code-generation training examples used for SemBound.
The code split contains 20,804 training examples in total, including web code, LeetCode function-completion records, contest-style programs, and targeted synthetic non-HumanEval samples.
}
\label{tab:sembound-code-source-breakdown}
\end{table}

\section{Code-Head Training Loss}
\label{app:code_loss}

This section defines the loss terms used for code-head training. All losses are computed only over valid supervised positions.

Let $\hat{y}_i$ denote the boundary logit at position $i$, $y_i \in \{0,1\}$ denote the boundary label, and $m_i^{b}$ denote the boundary loss mask. The masked boundary loss is
\[
\mathcal{L}_{\mathrm{bdry}}
=
\frac{
\sum_i m_i^{b} \, \ell_{\mathrm{BCE}}(\hat{y}_i, y_i)
}{
\sum_i m_i^{b}
}.
\]

Let $\hat{\mathbf{p}}_i$ denote the logits for program phase prediction, $p_i$ denote the phase label, and $m_i^{p}$ denote the phase supervision mask. The phase loss is
\[
\mathcal{L}_{\mathrm{phase}}
=
\frac{
\sum_i m_i^{p} \, \ell_{\mathrm{CE}}(\hat{\mathbf{p}}_i, p_i)
}{
\sum_i m_i^{p}
}.
\]

Let $\hat{t}_i$ denote the transition logit after position $i$, $t_i \in \{0,1\}$ denote the transition label, and $m_i^{t}$ denote the transition supervision mask. The transition loss is
\[
\mathcal{L}_{\mathrm{trans}}
=
\frac{
\sum_i m_i^{t} \, \ell_{\mathrm{BCE}}(\hat{t}_i, t_i)
}{
\sum_i m_i^{t}
}.
\]

Let $\hat{\mathbf{c}}_i$ denote the logits for boundary-type prediction, $c_i$ denote the boundary-type label, and $m_i^{c}$ denote the boundary-type supervision mask. This term is computed only where boundary-type labels are available:
\[
\mathcal{L}_{\mathrm{type}}
=
\frac{
\sum_i m_i^{c} \, \ell_{\mathrm{CE}}(\hat{\mathbf{c}}_i, c_i)
}{
\sum_i m_i^{c}
}.
\]

Let $q_i=\sigma(\hat{y}_i)$ denote the predicted boundary probability, and let $\rho$ denote the target boundary ratio. The rate regularization term controls the average predicted boundary density:
\[
\mathcal{L}_{\mathrm{rate}}
=
\left(
\frac{
\sum_i m_i^{b} q_i
}{
\sum_i m_i^{b}
}
-
\rho
\right)^2 .
\]

The final code-head objective is the weighted sum.

\section{Additional Details of Boundary Supervision}
\label{app:boundary-supervision}

For a training example, let $x_{1:n}$ denote the tokenized sequence. 
The boundary label $y_i \in \{0,1\}$ indicates whether the position after token $x_i$ closes a semantic unit. 
Although the definition of a semantic unit varies across domains, all supervision signals are converted into the same token-level boundary format:
\[
y_i =
\begin{cases}
1, & \text{if a semantic unit ends after } x_i, \\
0, & \text{otherwise}.
\end{cases}
\]
A loss mask $m_i$ specifies whether the boundary label at position $i$ is reliable and should contribute to training.

More generally, for a domain $d$, each example is represented as a sequence of semantic segments
\[
\mathcal{S}^{(d)} = \{s_1, s_2, \ldots, s_K\}.
\]
After tokenization, segment $s_k$ is aligned to a token interval $[a_k,b_k]$. 
The boundary label is assigned to the final token of each segment:
\[
y_i = \mathbb{I}\left(i \in \{b_1,b_2,\ldots,b_K\}\right),
\]
where $\mathbb{I}(\cdot)$ is the indicator function. 
Positions outside the supervised target region, padding positions, and positions with uncertain alignment are masked out by $m_i$.

\subsection{Natural Language Boundary Construction}
\label{app:nl-boundary-construction}

For natural-language data, discourse segmentation annotations are used to construct semantic units. 
Each discourse segment is treated as a locally coherent unit of meaning. 
The boundary supervision therefore corresponds to transitions between adjacent discourse units, rather than to punctuation marks or line breaks.

Formally, suppose a natural-language example is annotated as discourse segments
\[
\mathcal{U} = \{u_1,u_2,\ldots,u_K\}.
\]
The segments are concatenated in their original order and tokenized with the LLaDA tokenizer. 
If segment $u_k$ corresponds to token interval $[a_k,b_k]$, the label is defined as
\[
y_i^{\mathrm{nl}} = \mathbb{I}\left(i=b_k \text{ for some } k\right).
\]
Tokens inside a discourse segment receive non-boundary labels. 
This construction encourages the natural-language head to identify points where a coherent discourse unit has been completed.

The loss mask is enabled only for token positions that can be reliably recovered from the annotated discourse segments. 
Prompt tokens, padding tokens, and unstable alignment positions are excluded. 
Thus, the natural-language head learns from discourse-level transitions while keeping the final supervision format identical to the other domains.

\subsection{Mathematical Reasoning Boundary Construction}
\label{app:math-boundary-construction}

For mathematical reasoning data, semantic units are constructed from the solution rationale. 
A reasoning unit corresponds to a coherent step in the solution process, such as identifying quantitative relations, setting up equations, carrying out arithmetic or algebraic computation, comparing candidate answers, or producing the final answer.

Let $r$ denote the solution text of a mathematical example. 
The solution is first divided into candidate reasoning fragments. 
Short action-leading fragments are merged with the following equation or result fragment when they form a single reasoning action. 
This produces a sequence of reasoning units
\[
\mathcal{R} = \{r_1,r_2,\ldots,r_K\}.
\]
Each unit is intended to preserve a local reasoning transition, rather than a surface sentence boundary.

After tokenization, if reasoning unit $r_k$ corresponds to interval $[a_k,b_k]$, the mathematical boundary label is
\[
y_i^{\mathrm{math}} = \mathbb{I}\left(i=b_k \text{ for some } k\right).
\]
The boundary therefore indicates the completion of a reasoning step. 
This label construction is especially important for mathematical text, where a single sentence may contain multiple operations and a multi-line derivation may still belong to one coherent reasoning unit.

When a final answer anchor is available, the answer segment is treated as a special closure point of the reasoning process. 
This helps the mathematical head learn the transition from intermediate computation to the final conclusion. 
The resulting boundary sequence can also be converted into block lengths by measuring the token length of each reasoning unit, which is used by the oracle-style math data construction.

\subsection{Code Boundary Construction}
\label{app:code-boundary-construction}

For code generation, each example is organized as a function completion task. 
The conditioning context contains the task description, the function signature, and, when available, compact public-test information. 
The target region is the function body to be generated. 
Boundary supervision is applied mainly to this target code region, while the conditioning context provides the task information needed for generation.

Let $c$ denote the serialized code example, and let $\alpha$ be the character position where the target code region begins. 
A program analysis procedure produces a set of annotated spans
\[
\mathcal{P}=\{(l_j,r_j,z_j)\}_{j=1}^{J},
\]
where $[l_j,r_j]$ is a character span in the program and $z_j$ is its program phase label. 
The phase label describes the functional role of the span, such as input normalization, state initialization, condition checking, loop expansion, state update, result aggregation, or return emission.

Each character span is projected to a token interval after tokenization:
\[
\pi([l_j,r_j]) = [a_j,b_j].
\]
For every token position $i \in [a_j,b_j]$, the phase target is assigned as
\[
z_i = z_j.
\]
If incompatible phase assignments overlap at the same token position, the phase supervision at that position is masked. 
This keeps the structural supervision conservative when character spans and token boundaries do not align exactly.

Code boundaries are derived from structural closure points between program spans. 
Let $e_j$ denote a character-level boundary position. 
It is mapped to the token immediately before the boundary:
\[
i_j = \pi(e_j)-1.
\]
The boundary label is then assigned as
\[
y_i^{\mathrm{code}} = \mathbb{I}\left(i=i_j \text{ for some boundary } e_j\right).
\]
A positive label means that the program has completed a functional unit after token $x_i$.

When a boundary type is available, the mapped boundary position also receives a type label:
\[
c_{i_j}=c_j.
\]
The type label distinguishes different forms of structural closure, including statement completion, loop expansion, branch transition, local computation closure, and return emission. 
This supervision allows the code head to learn not only whether a boundary exists, but also what kind of program transition it represents.

The code training example maintains separate validity masks for boundary, phase, transition, and boundary type supervision. 
If a sequence is cut by the training length constraint, the final token of the truncated sequence is not treated as a reliable boundary. 
The boundary label and boundary type label at this position are cleared. 
This prevents artificial closure points from being introduced by truncation.




\section{Additional Experimental Details}
\label{app:additional_details}

\textbf{Hyperparameter Settings.}
For the main experiments, we set the generation length to 512 and the confidence threshold for dynamic sampling to 0.9. We evaluate default block sizes $B_0 \in \{16,32,64\}$ and use $B_0=32$ as the primary setting in the main comparison table.

\textbf{Task Settings.}
For GSM8K, MATH, and IFEval, we use 16 denoising steps. For the main HumanEval result on LLaDA-Instruct, we use 32 denoising steps. The main results on GSM8K, MATH, and IFEval use cache, while the main HumanEval result on LLaDA-Instruct does not use cache.

\end{document}